\documentclass[10pt,twocolumn,letterpaper]{article}

\usepackage{cvpr}              % To produce the CAMERA-READY version
\usepackage{graphicx}
\usepackage{amsmath}
\usepackage{amssymb}
\usepackage{booktabs}

\usepackage{enumitem}
\usepackage{multirow}

\usepackage[pagebackref,breaklinks,colorlinks]{hyperref}

\usepackage[capitalize]{cleveref}
\crefname{section}{Sec.}{Secs.}
\Crefname{section}{Section}{Sections}
\Crefname{table}{Table}{Tables}
\crefname{table}{Tab.}{Tabs.}

\def\cvprPaperID{6288} % *** Enter the CVPR Paper ID here
\def\confName{CVPR}
\def\confYear{2022}

\begin{document}

\title{Cross-Architecture Self-supervised Video Representation Learning}

% \author{
% Sheng Guo\textsuperscript{1} \quad Zihua Xiong\textsuperscript{1} \quad Yujie Zhong\textsuperscript{2} \quad Limin Wang\textsuperscript{3} \\
% \quad Xiaobo Guo\textsuperscript{1} \quad   Bing Han\textsuperscript{1} \quad Weilin Huang\textsuperscript{4}\\
% \textsuperscript{1}MYbank, Ant Group, China \\
% \textsuperscript{2}Meituan \\
% \textsuperscript{3} Nanjing University \\
% \textsuperscript{4}Alibaba Group \\
% }

\author{
    Sheng Guo$^1$\thanks{Equal contribution.}
    ~
    Zihua Xiong$^{1*}$
    ~
    Yujie Zhong$^2$
    ~
    Limin Wang$^3$
    ~
    Xiaobo Guo$^1$
    ~
    Bing Han$^1$
    ~
    {Weilin Huang}$^4$\thanks{Corresponding author.}\\
    \normalsize
    $^1$MYbank, Ant Group \quad 
    \normalsize
    $^2$Meituan Inc. \quad  \\
    \normalsize
    $^3$State Key Laboratory for Novel Software Technology, Nanjing University, China \quad 
     \normalsize
    $^4$Alibaba Group\\
    {\small \tt \{guosheng.guosheng,xiongzihua.xzh,jefflittleguo.gxb,hanbing.hanbing\}@mybank.cn
    }\\
    {\small \tt jaszhong@hotmail.com, lmwang@nju.edu.cn, weilin.hwl@alibaba-inc.com
    }\\
}
% \and
\maketitle

%%%%%%%%% ABSTRACT
\begin{abstract}
   In this paper, we present a new cross-architecture contrastive learning (CACL) framework for self-supervised video representation learning. CACL consists of a 3D CNN and a video transformer which are used in parallel to generate diverse positive pairs for contrastive learning. This allows the model to learn strong representations from such diverse yet meaningful pairs. 
   %which is the key to boost the performance.
%It also demonstrates that the two architectures of 3D CNN and video transformer can compensate strongly to each other. 
Furthermore,  we introduce a temporal self-supervised learning module able to predict an Edit distance explicitly between two video sequences in the temporal order.
%overcoming the key limitation of previous self-supervised temporal learning methods.
%This allows it to predict a difference in temporal order with arbitrary degree, overcoming the key limitation of previous self-supervised temporal learning methods, which often estimate a very limited degree of temporal difference (e.g., two videos are in a same order or not). 
This enables the model to learn a rich temporal representation that compensates strongly to the video-level representation learned by the CACL. 
%In addition, the temporal prediction module is integrated into the cross-architecture framework to perform an end-to-end training.
We evaluate our method on the tasks of video retrieval and action recognition on UCF101 and HMDB51 datasets, where our method achieves excellent performance, surpassing the state-of-the-art methods such as VideoMoCo \cite{Pan2021VideoMoCoCV} and MoCo+BE \cite{wang2021removing} by a large margin. The code
is made available at \url{https://github.com/guoshengcv/CACL}.
\end{abstract}

%%%%%%%%% BODY TEXT
\vspace{-6mm}
\section{Introduction}
\label{sec:intro}

Video representation learning is a fundamental task for video understanding, as it plays an important role on various tasks, such as action recognition~\cite{Simonyan2014,tran2015learning,tsn,tdn,zhang2020action}, video retrieval~\cite{zhang2020,mmn}, and video temporal detection~\cite{ssn,BMN,rtd}. Recent efforts have been devoted to improving its performance by using deep neural networks in a supervised learning manner, which often requires a large-scale video dataset with very expensive human annotations, such as Sports-1M \cite{KarpathyCVPR14}, Kinetics \cite{kay2017kinetics}, HACS \cite{zhao2019hacs}, and MultiSports~\cite{multisports}. 
The large annotation cost inevitably limits the potential of deep networks on learning video representation. Therefore, it is of great importance to improve this task by leveraging  unlabeled videos which are easily accessible at a large scale.

\begin{figure*}[t]
\centering
  \includegraphics[scale=0.27]{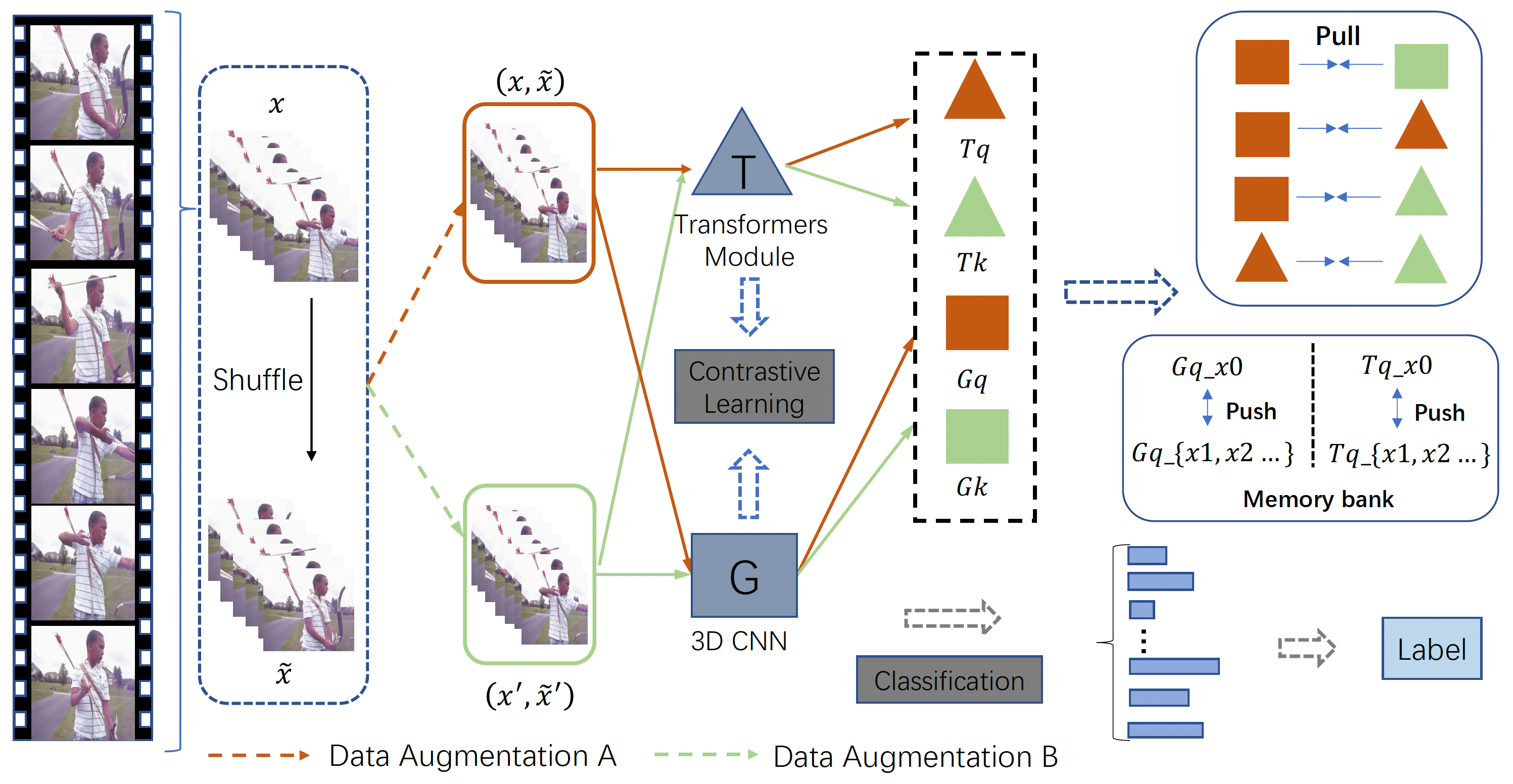}
  \vspace{-3mm}
 \caption{Illustration of the proposed CACL framework. 
 (i) Given a video clip $x$, we implement random shuffle on temporal dimension to generate a shuffled version $\widetilde x$, with random data augmentations applied.  (ii) Then the original clip and the shuffled one are concatenated and passed to a transformer encoder, generating a transformer-based video representation.  At the same time,  $x$ and $\widetilde x$ are computed by the 3DCNN encoder separately to generate a 3D video representation.  In total, the two different encoders output four different representations of the input video, resulting in four positive pairs for constractive learning.  (iii) Furthermore, the two 3D representations are concatenated, and are used for self-supervised temporal learning by predicting an Edit distance, which can be reformulated as a classification task.  We demonstrate that the temporal representation learned by our method can compensate strongly to our video-level representation (by CACL).}
 %and the two tasks can work collaboratively in our multi-task learning framework.}
 %For the contrastive learning, we adopted MoCo to maximize the similarities of positive pairs constructed, and minimize the similarities between positive samples and negative ones popped from a memory bank. 
 %Shuffle degree prediction task is formulated as a classification problem.}
\label{fig:learning_scheme}
% \vspace{-4mm}
\end{figure*}

Recently, self-supervised learning has made significant progress on learning strong image representation, by constructing various supervised learning signals from images themselves. It has also been extended to video domain, where contrastive learning has been widely-applied. For example, in recent works such as~\cite{Pan2021VideoMoCoCV, wang2020self, tao2020self} contrastive learning was introduced to capture the discrimination between two video instances, which enables it to learn an invariant representation within each video instance.
However, the contrastive learning mainly focuses on learning a global spatio-temporal representation of videos in these approaches, while it is difficult to capture meaningful temporal details which often provide important cues for discriminating different video instances, e.g., human actions.
Therefore, different from learning image representation, modelling temporal information is critical to video representation. 
In this work, we present a new self-supervised video representation method able to perform both video-level contrastive learning and  temporal modelling \textit{simultaneously} in a unique framework, as shown in Figure \ref{fig:learning_scheme}.

The supervised signal of learning temporal information can be created by exploring the sequential nature of videos, allowing for performing self-supervised learning. 
Recent methods, such as pace predictions~\cite{Wang2020, Yang2020} and playback speeds perception~\cite{Yao2020,chen2021rspnet}, followed this line of research by creating a pretext task that implements self-supervised temporal predictions.
%take the usage of such video sequential nature.
%
In this work, we introduce a new self-supervised temporal learning by predicting an approximate Edit distance between a video (i.e. a sequence of frames) and its temporal shuffle. This allows us to \textit{explicitly} measure the degree of temporal difference quantitatively in Edit distance, setting it apart from the existing self-supervised methods which are often limited to estimate a rough difference of two videos in the temporal domain. For example, they often created a pretext task to predict whether two video sequences are in the same pace or playback speeds, but ignore details in such temporal difference. 

While most self-supervised contrastive learning methods generate positive pairs using various data augmentations which provide different views of an instance, we develop a new method able to learn stronger representation from diverse architectures via contrastive learning.
The family of 3D CNNs has achieved remarkable performance in various video tasks, including C3D~\cite{tran2015learning}, R3D~\cite{hara2018can}, R(2+1)D~\cite{tran2018closer}, etc. 
They are capable of capturing local dependencies in the temporal domain due to the intrinsic property (i.e., convolutions) of CNNs. 
But the effective receptive fields of CNNs might limit their ability to modelling long-range dependencies.  
On the other hand, such long-range dependencies can be naturally captured by the transformer architecture~\cite{vaswani2017attention} using a self-attention mechanism, where each token is able to learn an attention to the whole sequence, and thus encodes meaningful context information into video representations.
%It makes transformers good encoders for video representations. %
Moreover, the inductive biases of CNNs may limit their performance when trained on sufficiently large data, while this limitation may not happen in the transformers due to the dynamic weighting of self-attention~\cite{dosovitskiy2021an}.

We argue that modelling both local and global dependencies are essential for video understanding; and the inductive biases of CNNs and the capacity of transformers can compensate strongly to each other. 
In this work, we present a new cross-architecture contrastive learning (CACL) framework for self-supervised video representation  learning. 
Our CACL is able to learn from diverse yet more meaningful constrastive pairs generated by a 3D CNN and a video transformer. 
We demonstrate that a video transformer can strongly enhance the video representation generated by a 3D CNN. It produces rich high-level contextual features and encourages the 3D CNN to capture more detailed information. This allows the two architectures to work collaboratively, which is the key to boost the performance.
%
%To summarize,  we present a new self-supervised framework for video representation learning, as illustrated in Figure \ref{fig:learning_scheme}. It is able to learn meaningful video-level global features with the enhanced temporal information, resulting in a strong global-temporal video representation. 
Our main contributions can be summarized as follows:

%\begin{itemize}%[leftmargin=10pt,topsep=0pt]

%\item[--] 
-- We design a new cross-architecture contrastive learning (CACL) framework for self-supervised video representation learning. 
CACL uses a  3DCNN and a Transformer to collaboratively generate diverse yet meaningful positive pairs, which allow for more effective contrastive representation learning.

%\item[--] 
-- We introduce a new self-supervised temporal learning method by explicitly measuring an edit distance between a video and its temporal self-shuffle. This helps to learn rich temporal information complementary to the representation learned from our CACL.

%\item[--] 
-- We verify our method on two downstream video tasks: action recognition and video retrieval. Experimental results on UCF101~\cite{soomro2012ucf101} and HMDB51~\cite{kuehne2011hmdb} show that the proposed CACL can significantly outperform the state-of-the-art methods, such as VideoMoCo \cite{Pan2021VideoMoCoCV} and MoCo+BE \cite{wang2021removing}.

%\end{itemize}

%------------------------------------------------------------------------
\section{Related Work}

%Recently， there are wide investigations in self-supervised representation learning in both image domain and video domain. 

%The principle of self-supervised representation learning is to define an annotation-free proxy task that encourages a model to learn semantic and generic video features from unlabeled data, which can benefit various downstream tasks. 

%Transformers show its effectiveness in static image tasks, and has extend to video tasks recently. 
In this section, we briefly introduce two groups of self-supervised methods for video representation learning: contrastive learning methods and pretext-task approaches. Then we briefly review recent studies on applying transformers in the related video tasks.

\paragraph{Contrastive learning methods.} Self-supervised contrastive learning has received extensive attention. The common approach is to maintain a relative consistency between the representations of an instance and its augmented view. It has achieved remarkable success in image-level representation learning,  such as SimCLR~\cite{chen2020simple} and MoCo~\cite{he2020momentum}, and recent studies extend the contrastive learning paradigm to video applications. %
For example, VideoMoCo~\cite{Pan2021VideoMoCoCV} extends image-based MoCo framework to video representation learning by enhancing the temporal robustness of the encoder, and at the same time, modeling the temporal decay of the keys. 
In~\cite{Wang2021EnhancingUV}, Wang \etal proposed DSM able to construct positive and negative sample pairs, which alleviate the negative impact of the scene and the motion coupling problem. 
TCLR~\cite{dave2021tclr} was developed by introducing two different temporal contrastive losses, in order to learn temporally distinct features across the video. It is combined with the vanilla instance contrastive loss, leading to an increase in the temporal diversity of the learned features. 
In~\cite{qian2021spatiotemporal}, CVRL was introduced by studying various data augmentations for video self-supervised learning. Then a temporally consistent spatial augmentation and a sampling-based temporal augmentation methods were proposed. These approaches perform both spatial-temporal data augmentations to construct positive samples for contrastive learning. In this work, we generate the positive sample pairs from two different network architectures able to learn two diverse video embeddings that focus on learning 3D local video details or long-range temporal dependencies. We demonstrate that the transformer architecture can provide the positive samples with more diversity for the 3D CNN.

\paragraph{Pretext task based approaches.} It is intuitive to exploit temporal dimension to generates pseudo training labels without human annotations. Many recent works rely on investigating the speed property of videos, such as pace predictions~\cite{Wang2020, Yang2020}, playback speeds perception~\cite{Yao2020,chen2021rspnet}, and speediness of moving objects~\cite{benaim2020speednet}. 
On the other hand, some recent studies focus on learning from video temporal order e.g., by verifying the correctness of temporal order of a video. In \cite{misra2016shuffle} Misra \etal created a sequential verification task by comparing a video sequence with its shuffled version, and then OPN~\cite{opn} extends the sequential verification to order prediction. This allows it to generate richer learning signals that enables the model to learn more temporal details. In~\cite{xu2019self}, VCOP was proposed to learn the spatio-temporal representation of the video by predicting the order of shuffled clips from a video. However, the degree of a video temporal shuffle has not  been considered among all these methods. In this work, we propose to use the degree of temporal order as supervision signals which enable the model to learn more detailed temporal information.

\paragraph{Transformers.} The transformer~\cite{vaswani2017attention} has recently achieved remarkable performance in various computer vision tasks.
%, which traditionally depends on deep ConvNets.  
For example, ViT~\cite{dosovitskiy2021an} and DeiT~\cite{Touvron2021TrainingDI} attempted to perform image recognition by using transformers. They split an image into a set of image patches, generating a sequence of patch embeddings which are used as an input to the transformer.Both methods achieve excellent results compared to state-of-the-art convolutional networks. Besides, Carion \etal proposed DETR~\cite{Carion2020EndtoEndOD}, where a direct set prediction approach is equipped with a transformer encoder-decoder architecture, yielding excellent results on par with the well-established CNN-based object detector.

With the sequential nature of videos, it is natural to apply the transformer for better modelling video related tasks. Recently, Wang \etal developed VisTR~\cite{Wang2020EndtoEndVI} which is a simpler and faster video instance segmentation framework based on transformers.
In~\cite{Neimark2021VideoTN}, VTN was presented, consisting of a transformer module to process spatial-temporal information. This started a new line of research in video recognition domain. 
Our CACL further explore the potential of transformer in modeling long-range dependencies, and the architecture discrepancy between the transformer and convolutional networks. This allows us to build a transformer video encoder that provides rich compensatory features for 3D CNN in contrastive learning.

%-------------------------------------------------------------------------------------
\section{Methodology}

We tackle video representation learning problem in a self-supervised manner. In this section, we first present an overall framework of the proposed method. Then we describe details of the proposed contrastive learning method, and our self-supervised temporal learning based on a prediction of frame-level shuffle degree.

\subsection{Overall Framework}

Our framework consists of two pathways including a transformer video encoder and a 3D CNN video encoder. The self-supervised learning signals are computed from two tasks: clip-level contrastive learning and frame-level temporal prediction.

\paragraph{3D CNN video encoder.} 
In this work , 3D CNNs (such as C3D~\cite{tran2015learning}, R3D~\cite{hara2018can} and R(2+1)D~\cite{tran2018closer}) are adopted as the main video encoder, which are also used for inference. Notice that any other 3D CNN architectures can also be applied in our framework. The output features of original clip and shuffled clip are concatenated and then passed through the contrast head and the classification head. Both heads are fully connected feed-forward networks.

\begin{figure}[t]
\centering
  \rotatebox{90}{\includegraphics[scale=0.4]{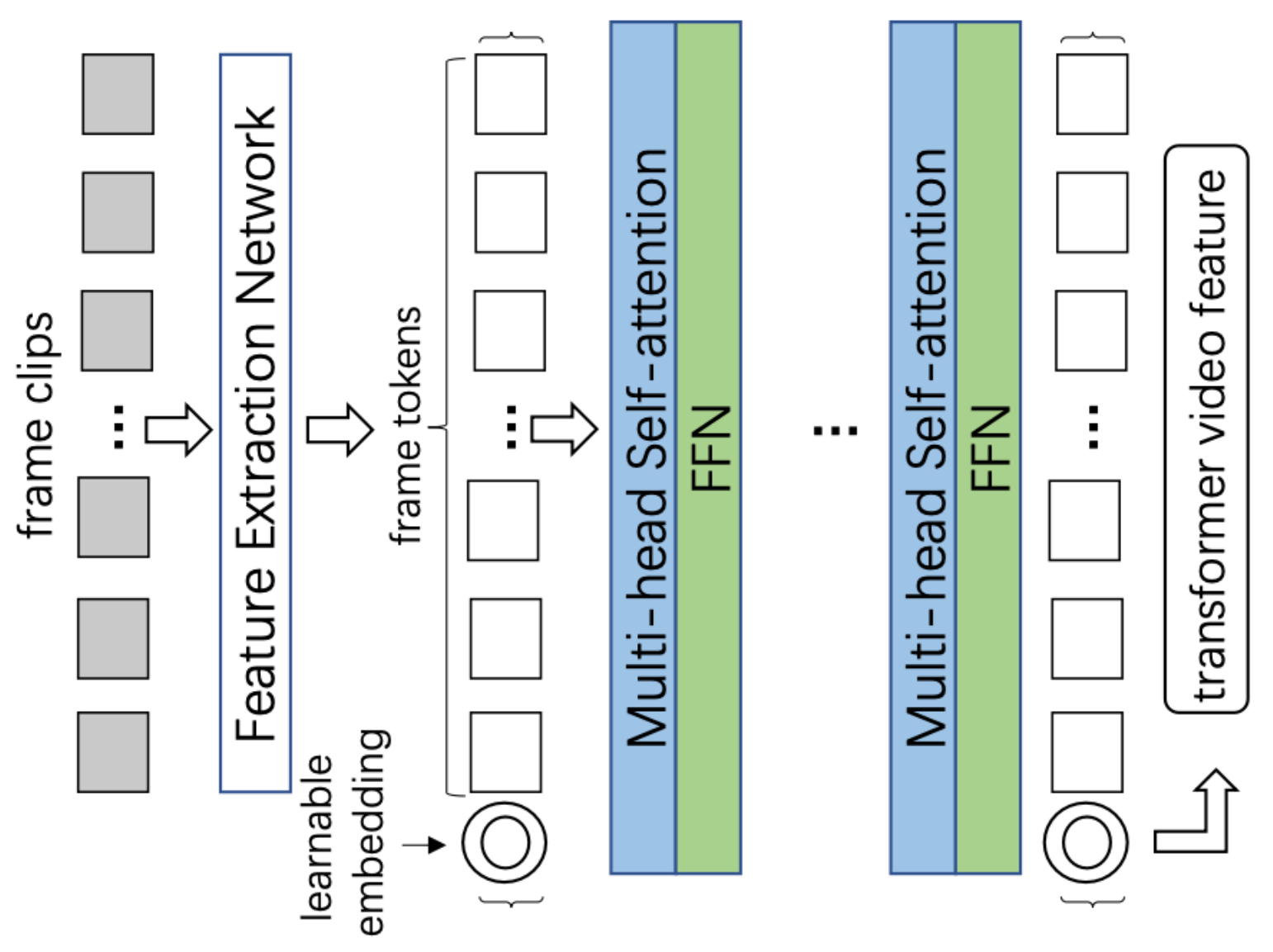}}
  \vspace{-3mm}
 \caption{Transformer video encoder architecture.}
\label{fig:transformer}
% \vspace{-4mm}
\end{figure}

\paragraph{Transformer video encoder.} The transformer encoder consists of a 2D CNN and a transformer architecture, which is motivated by VTN~\cite{Neimark2021VideoTN}. 
The pipeline is illustrated in Figure \ref{fig:transformer}. Firstly, individual image frames of a video clip are computed through a 2D CNN which performs feature extraction to obtain a sequence of frame-level tokens. The output CNN features are then projected to 768-D frame tokens through a fully connected layer. Then the frame tokens are concatenated sequentially in a temporal order, and a learnable embedding is prepended to the sequence of frame tokens. Finally, a 6-layer 6-head transformer model takes the clip-level feature sequence as input, and the output of learnable embedding is used as video representation.
Notably,  the feature extraction network is ResNet50~\cite{he2016deep} pretrained with video frames from UCF101 train set, by using self-supervised method MoCo~\cite{he2020momentum}. The pretrain details will be illustrated in \ref{subsec:Implementation}, and its weights are frozen during self-supervised video representation learning.

%\paragraph{Training samples.} Firstly, given a video instance, we extract one clip $x$ by starting from a random timestamp with an equal sampling rate. Then we perform random shuffle on temporal dimension to obtain a shuffled video clip $\widetilde x$, with random data augmentation. For a transformer video encoder, we concatenate $x$ and $\widetilde x$, and pass it to the transformer video encoder to generate a transformer-based video representation. For a 3D CNN video encoder, we pass $x$ and $\widetilde x$ to a 3D CNN separately. Then the output features are concatenated, and passed to a contrast head and a classification head to perform two learning tasks jointly.

%\subsection{Cross-Architecture Noise Contrastive Estimation}

\subsection{Cross-Architecture Contrastive learning}

% Contrastive learning in a self-supervised manner has shown great potential and 
% achieved comparable results with supervised visual representation learning recently.
The goal of our self-supervised contrastive learning is to maximize the similarity between the video clips with same context, while minimizing the similarity of clips sampled from different videos.
Different from previous contrastive learning methods,our CACL leverages cross-architecture contrastive learning signals to better capture both local and  long-range dependencies jointly. 

\paragraph{Construction of positive pairs.}
The fundamental problem of contrastive learning lies in the design of positive and negative samples. Previous works on self-supervised contrastive learning often utilize various data augmentations to generate different transformed versions of a particular instance (e.g. different clips of the same video) which forms the positive pairs. 
In this work, we enrich the positive pairs from two perspectives: embedding level (using different architectures) and data level.
From the perspective of architecture, CACL takes the advantage of a 3D CNN and a transformer. Given an input video clip, each of them produces a video representation separately, which doubles the number of positive samples comparing to previous methods.
For the data level, we apply a random shuffle on temporal dimension on the original clip $x$ and obtain a shuffled video clip $\widetilde x$. The two instances are latter concatenated.
 As illustrated in Figure \ref{fig:learning_scheme}, we maximize the similarities of four positive pairs generated from each video clip, by using different data augmentations and different encoders. We denote different data augmentations as $q,k$, a transformer encoder as $T$, a 3D CNN encoder as $G$, then we can generate four feature representations $Gq, Gk, Tq,Tk$, for a video clip.

\paragraph{Negative pairs.}
Clips from different videos are considered as negative samples. We further enhance the contrastive learning by using a momentum encoder and a memory dictionary queue motivated by MoCo~\cite{he2020momentum}, providing more meaningful negative samples for improving the performance of contrastive learning.

\paragraph{Data augmentations.}
Data augmentations are performed in both spatial and temporal domains on the input video clips.
Note that the spatial augmentation is performed consistently across all frames within a clip. 
Therefore, we maximize three kinds of  similarities: (1) the similarity between the clips computed by the same network but performed different data augmentations; (2) the similarity between the clips with the same data augmentation but computed by different networks; (3) the similarity  between the clips using different networks with different data augmentations.
%In particular, we can reinforce them with various data augmentations performed in both spatial and temporal domains,  and different modality video encoders, such as a 3D CNN video encoder and a transformer video encoder.

\paragraph{Loss function.}
Formally, we consider a randomly sampled mini-batch consisting of $N$ different video instances, and then we sample one  clip from each video starting a random timestamp with an equal sampling rate. This results in a total of $N$ clips ($C$) in a mini-batch.
We randomly shuffle the order of each clip, yielding a new  set of $N$ clips ($C_{s}$). 
Then each clip and its shuffled version are concatenated, and are further processed with data augmentations. This generates two concatenated video clips only having different data augmentations. The generated clips are separately processed by different video encoders:   a  3D-CNN based video encoder and a transformer-based encoder. Therefore, we generate four clip-level video representation for each video instance: $Gq_{i}, Gk_{i}, Tq_{i},Tk_{i}$, which are used to construct positive pairs during constractive learning.
%
%This pair of clips $(C,C_{s})$ has been concatenated and enhanced with data augmentation  to get two  clips, $(C^{t1},C^{t1}_{s})$ and $(C^{t2},C^{t2}_{s})$.  Each of the transformed clip is then separately  passed through a  3D-CNN based video encoder  and  a transformer module based video encoder.   Hence, for each video-instance $i$ we get four  clip representations: $Gq_{i}, Gk_{i}, Tq_{i},Tk_{i}$. 
%
We leverage the idea of instance discrimination using InfoNCE~\cite{oord2018representation}  based contrastive loss.
% $(Gq_{i}, Gk_{i}), (Tq_{i},Tk_{i}), (Gq_{i}, Tq_{i}),(Gq_{i},Tk_{i})$
\begin{small}
\begin{align}
    \label{eqn:nce}
    \mathcal{L_{\text{NCE}}} &= - \frac{1}{n} \sum_{i=1}^n \log\frac{e^{cont(Gq_i,Gk_i)}}{\sum_{j=1}^n e^{cont(Gq_i,Gk_j)}}, %\\ [8pt]
\end{align}
\end{small}

\begin{small}
\begin{align}
    \label{eqn:conut}
    cont(Gq, Gk) &= \tau * \frac{G(q)}{\| G(k) \|} \cdot \frac{G(q)}{\| G(K) \|},
\end{align}
\end{small}

\noindent where $cont(Gq,Gk)$ is the similarity metric between two vectors. $Gq$ and $Gk$ are two feature representations. $\tau$ is  an  adjustable parameter.
In this work, we extend the constrastive learning for video representation learning as:

\begin{equation}
\begin{split}
    \label{eqn:nce_pos}
 S^i_\text{pos}= e^{cont(Gq_i,Gk_i)}  + e^{cont(Gq_i,Tq_i)}  \\
     + e^{cont(Gq_i,Tk_i)} + e^{cont(Tq_i,Tk_i)},
\end{split}
\end{equation}

\begin{small}
\begin{align}
    \label{eqn:batch_nce}
    \mathcal{L_{\text{NCE}}} &= - \frac{1}{n} \sum_{i=1}^n \log\frac{S_\text{pos}^{i}}{S_\text{pos}^{i} + \sum_{j=1}^m e^{cont(Gq_i,N_j)}},
%\label{eqn:nce}
\end{align}
\end{small}

\noindent where $N_j$ is a negative sample from a memory dictionary queue with a queue size of $m$.
As shown in Equation~\ref{eqn:batch_nce}, our CACL is able to generate more positive pairs than the standard contrastive learning.

\subsection{Temporal Prediction with Edit Distance}

We aim to learn video representations that are sensitive to temporal details. To this end, we attempt to train the network by explicitly predicting the temporal difference between a video clip and its shuffled version. 
We assume that such a temporal prediction task requires both motion and appearance cues. This enables the model to learn meaningful temporal details which in turn benefit downstream tasks. 
In this work, we propose to use a Minimum Edit Distance (MED) to measure the degree of temporal difference between a video clip and its shuffled version. 

MED was originally introduced in~\cite{levenshtein1966binary} by  Vladimir Levenshtein in 1965, and it provides a way to measure the dissimilarity between two strings (e.g., words), by counting the minimum number of operations required to transform one string into the other. Mathematically, the Levenshtein distance between two strings $a$, $b$ is given by $\text{lev}_{a,b}(\lvert a \rvert, \lvert b \rvert)$ where

\begin{align}
    \begin{scriptsize}
    \label{eqn:nce}
        \text{lev}_{a,b}\left ( i,j \right ) = \begin{cases}
        \text{max}(i,j)  \qquad \text{ if } \text{min}(i,j)= 0,\\ 
        \text{min}\begin{cases}
        \text{lev}_{a,b}(i-1,j)+1 & \\
        \text{lev}_{a,b}(i,j-1)+1 & \\ 
        \text{lev}_{a,b}(i-1,j-1)+I(a_{i}\neq b_{j})&
        \end{cases}  \text{ otherwise. } 
        \end{cases}
    \end{scriptsize}
\end{align}

\noindent where $I(a_{i}\neq b_{j})$ is an indicator function which is equal to 0 when $a_{i}= b_{j}$, and is 1 otherwise.
%For a video instance, we extract one clip starting from a random timestamp with the same  sampling rate. We randomly shuffle the original clip to obtain a shuffled one, and then pass them through a 3D CNN video encoder separately, and finally concatenate the two features, which are used as inputs of the classification head for temporal prediction. 
In this work, the shuffle degree prediction task is formulated as a classification problem, and the 3D CNN model $f(x, \widetilde x)$ is trained with a cross entropy loss.
Given a video clip $x$ and its shuffled version $\widetilde x$, we have:

\begin{align}
	\small
    \label{eqn:nce}
    \mathcal{L_{\text{cls}}} &= - \sum_{i=1}^m y_i \log\frac{e^{h_i}}{\sum_{j=1}^m e^{h_j}}, h=f(x, \widetilde x), %\\ [8pt]
\end{align}

\noindent where $m$ is the number of all the shuffle degree candidates.

\paragraph{Uniform shuffle-degree sampling.}
Given a video clip with 16 frames, we random shuffle it and compute a MED between the original clip and the shuffled one. Interestingly, we found that in our case the MED is a discrete integer ranging from 0 to 16 (except 1), which allows us to reformulate the regression problem of MED prediction into a classification task. However, the distribution of such discrete integers is not uniform, which may result in a unbalanced classification, making the training process unstable. Technically, we first randomly sample a MED number from  a uniform  distribution, and then randomly shuffle the video clip until it satisfies the sampled MED number. This operation allows us to well balance label distribution in classification, which is important to temporal modelling and joint learning.

\paragraph{Compared with other shuffle\&learn methods.} In contrast to earlier methods such as Shuffle\&Learn~\cite{misra2016shuffle}, OPN~\cite{opn} and VCOP~\cite{xu2019self}. Our method focuses on degree perception rather than order prediction/verification, which naturally leads to the following characteristics. It can learn more meaningful temporal information by increasing the number of frames, while previous methods are commonly limited to using a very small number of frames, e.g., 3 frames/clips. Because the number of orders overgrows with the increase in the number of frames/clips. Our method can capture more detailed and meaningful difference between video clips, allowing for learning richer temporal features.

%---------------------------------------------------------------------------------
\section{Experiments and Results}

\subsection{Implementation Details}
\label{subsec:Implementation}
\paragraph{Datasets.} UCF101~\cite{soomro2012ucf101} is an action recognition dataset which consists of 13,320 videos from 101 realistic action categories on YouTube. Following previous works~\cite{xu2019self,benaim2020speednet}, we use training split 1 for our self-supervised learning, training/testing split 1 for evaluating video retrieval task, and all training/testing splits for action recognition. HMDB51~\cite{kuehne2011hmdb} dataset consists of 6,849 clips from 51 action classes, collected from a variety of sources such as digitized movies and YouTube. Following~\cite{xu2019self,benaim2020speednet}, we use training/testing split 1 to perform video retrieval, and all training/testing splits for action recognition.

\paragraph{Pre-training feature extraction network.} We train our feature extraction network applied in video transformer by using self-supervised method MoCo~\cite{he2020momentum}. We follow the training policy with the same practice as in MoCo. 
To avoid introducing additional data, we form our training dataset by sampling frames from UCF101 training split 1. Specifically, we sample frames from a video with 10 intervals in order to enlarge inter-frame differentiation. 
With around 1.8M total frames in UCF101 training split 1, we can collect a training dataset consisting of approximately 180k frames.
%We form a train dataset with 180k images from UCF101 training split 1.

\paragraph{Self-supervised learning.} We train the network with our CACL framework for 300 epochs, and adopt SGD as our optimizer with a momentum of 0.9 and a weight decay of 5e-4. We set a batch size as 64. A learning rate is initialized as 0.001 with a cosine learning rate schedule used. We sample 16 frames with 2 intervals from a video. Video frames are first resized to 128 $\times$ 171, and then randomly cropped to 112 $\times$ 112. Meanwhile, we apply consistent data augmentations, such as horizontal flipping, color jittering, and random gaussian blur to all frames within a same video clip.

% \paragraph{Network Architecture.} For video encoder, 3D CNN networks such as C3D~\cite{tran2015learning}, R3D~\cite{hara2018can} and R(2+1)D~\cite{tran2018closer} are used. ViT is adapted to video transformer encoder, we illustrate video transformer encoder architecture following. ResNet50 is used as 2D image feature extraction network, and pre-trained by self-supervised method using video frames from UCF101. Transformer encoder consist of 6 attention blocks, each block has multi-head self-attention and feed-forward network, the number of multi-head is 6. The frame tokens and learnable embedding has 768-dims, and the output of learnable embedding serves as transformer video feature.

\paragraph{Supervised fine-tuning and evaluation.} After self-supervised learning, we transfer the weights of 3D CNN encoder network to action recognition. We fine-tune it on each dataset for 150 epochs. The momentum is set as 0.9, and the batch size is 128. The learning rate is initialized as 0.1 with a cosine learning rate schedule applied. For evaluation, by following common practice, the final result of a video is the average of the results of 10 clips uniformly sampled from the video. As the self-supervised training, we sample 16 frames with 2 intervals to form a clip.

\subsection{Ablation Study}

\begin{table}[!t]
%\footnotesize
\small
\centering
% \begin{tabular}{p{2.5cm}p{0.4cm}p{0.4cm}p{0.4cm}p{0.4cm}p{0.4cm}p{0.4cm}}
\begin{tabular}{l c c c c c}
\toprule
\bf Method & \bf 1 & \bf 5 & \bf 10  & \bf 20 & \bf 50\\
\midrule
VCOP (C3D)~\cite{xu2019self} & 12.5 & 29.0 & 39.0 & 50.6 &  66.9 \\
PRP (C3D)~\cite{Yao2020} & 23.2 & 38.1 & 46.0 & 55.7 &  68.4 \\
\midrule
SDP$^{\ast}$ & 17.2 & 34.6 & 45.7 & 57.8 &  74.3 \\
SDP & 22.1 & 40.6 & 50.3 & 61.2 &  75.5 \\
\midrule
V-MoCo & 30.3 & 47.4 & 57.0 & 66.3 & 78.9 \\
SDP+V-MoCo & 36.2 &55.0 &64.5 &72.8 & 83.2 \\
V-MoCo+T & 39.1 &53.3 &61.8 &72.2 & 85.0 \\
\midrule
SDP+V-MoCo+T & 43.2 &61.1 &69.9 &78.2 & 88.2 \\
\bottomrule
\end{tabular}
\vspace{-3mm}
\caption{ Ablation Studies on video retrieval  (on UCF101). SDP refers to shuffle degree prediction, SDP$^{\ast}$ means SDP without uniform sampling, and V-MoCo is a video-level contrastive learning by simply implementing MoCo with a 3D CNN encoder. $+T$ refers to a joint video transformer encoder.}
\label{tab:effectiveness_of_shuffle_degree}
\end{table}

% \begin{table}[h]
\begin{table}[!t]
%\small
\footnotesize
\centering
% \begin{tabular}{p{3.5cm}p{0.4cm}p{0.4cm}p{0.4cm}p{0.4cm}p{0.4cm}}
% \begin{tabular}{l p{0.4cm}<{\centering} p{0.4cm}<{\centering} p{0.4cm}<{\centering} p{0.4cm}<{\centering} p{0.4cm}<{\centering}}
\begin{tabular}[t]{l c c c c c}
\toprule
\bf positive pairs & \bf 1 & \bf 5 & \bf 10  & \bf 20 & \bf 50\\
\midrule
\tiny{($Gq,Gk$)} & 36.2 &55.0 &64.5 &72.8 & 83.2 \\
\tiny{($Gq,Gk$),($Gq,Tq$)} & 37.8 & 54.9 & 64.2 & 73.8 & 84.4 \\
\tiny{($Gq,Gk$),($Gq,Tq$),($Gq,Tk$)} & 39.3 &58.2 &67.0 &75.8 & 85.6 \\
\tiny{($Gq,Gk$),($Gq,Tq$),($Gq,Tk$),($Tq,Tk$)} & 43.2 &61.1 &69.9 &78.2 & 88.2 \\
\bottomrule
\end{tabular}
\vspace{-3mm}
\caption{Video retrieval with different positive pairs.}
\label{tab:effectiveness_of_video_transformer}
\end{table}

% \begin{table}[!h]
\begin{table}[!t]
\small
\centering
% \begin{tabular}{p{1.0cm}p{1.2cm}p{1.2cm}p{1.2cm}p{1.2cm}}
\begin{tabular}[t]{l c c c c}
\toprule
\bf Net & \bf ($Gq,Gk$) & \bf ($Gq,Tq$) & \bf ($Gq,Tk$)  & \bf ($Tq,Tk$) \\
\midrule
C3D & 0.8383 & 0.7844 & 0.6487 & 0.8534 \\
R(2+1)D & 0.8174 & 0.7773 & 0.6374 & 0.843 \\
R3D & 0.8128 & 0.7639 & 0.628 & 0.8427 \\
\bottomrule
\end{tabular}
\vspace{-3mm}
\caption{Average similarities on UCF101.}
\label{tab:positive_pairs_analysis}
\end{table}

\begin{table*}[ht]
%\scriptsize
\small
\centering
\begin{tabular}{p{3cm}c|ccccc|ccccc}
\toprule
    \multirow{2}*{\bf Method} & \multirow{2}*{\bf Net}
    & \multicolumn{5}{c|}{\bf UCF101} & \multicolumn{5}{c}{\bf HMDB51}\\
    \cline{3-12} 
    & & \bf 1 & \bf 5 & \bf 10  & \bf 20 & \bf 50 & \bf 1 & \bf 5 & \bf 10  & \bf 20 & \bf 50 \\
\midrule
% Jigsaw~\cite{noroozi2016unsupervised} & CFN  &  19.7 & 28.5 & 33.5 & 40.0 & 49.4 &-&-&-&-&- \\
% OPN~\cite{opn} & OPN  & 19.9 & 28.7 &  34.0 &  40.6 &  51.6 &-&-&-&-&-\\
% \midrule
VCOP~\cite{xu2019self} & C3D & 12.5 & 29.0 & 39.0 & 50.6 &  66.9 & 7.4 & 22.6 & 34.4 & 48.5 &  70.1 \\
Pace Pred~\cite{Wang2020} & C3D & 31.9 & 49.7 & 59.2 & 68.9 & 80.2 & 12.5 & 32.2 & 45.4 & 61.0 & 80.7 \\
DSM~\cite{Wang2021EnhancingUV} & C3D & 16.8 &33.4 &43.4 &54.6 & 70.7 & 8.2 &25.9 &38.1 &52.0 & 75.0 \\
RSPNet~\cite{chen2021rspnet} &C3D&36.0& 56.7& 66.5& 76.3& 87.7 &-&-&-&-&-\\
\textbf{CACL} & C3D &  \textbf{43.2} &  \textbf{61.1} &  \textbf{69.9} &  \textbf{78.2} &  \textbf{88.2} &  \textbf{17.3} &  \textbf{40.2} &  \textbf{53.8} &  \textbf{68.0} &  \textbf{85.2} \\
\textbf{CACL$^{\ast}$} &C3D&44.2& 63.1& 71.9& 80.4& 89.4 &-&-&-&-&-\\
\midrule
VCOP~\cite{xu2019self} & R3D & 14.1 & 30.3 & 40.0 & 51.1 & 66.5 & 7.6 & 22.9 & 34.4 & 48.8 & 68.9\\
Pace Pred~\cite{Wang2020} &R3D & 23.8 & 38.1 & 46.4 & 56.6 & 69.8 & 9.6 & 26.9 & 41.1 & 56.1 & 76.5 \\
\textbf{CACL} & R3D &  \textbf{41.1} &  \textbf{59.2} &  \textbf{67.3} &  \textbf{75.2} &  \textbf{85.9} &  \textbf{17.6} &  \textbf{36.7} &  \textbf{48.4} &  \textbf{63.0} &  \textbf{81.8} \\
\midrule
VCOP~\cite{xu2019self} & R(2+1)D & 10.7 & 25.9 & 35.4 & 47.3 & 63.9 & 5.7 & 19.5 & 30.7 & 45.8 & 67.0 \\
Pace Pred~\cite{Wang2020} & R(2+1)D & 25.6 & 42.7 & 51.3 & 61.3 & 74.0 & 12.9 & 31.6 & 43.2 & 58.0 & 77.1 \\
\textbf{CACL} & R(2+1)D &  \textbf{41.5} &  \textbf{59.7} &  \textbf{68.4} &  \textbf{77.6} &  \textbf{87.8} &  \textbf{16.4} &  \textbf{38.0} &  \textbf{49.6} &  \textbf{63.4} &  \textbf{81.1} \\
\bottomrule
\end{tabular}
\vspace{-3mm}
\caption{{\bf Recall-at-topK (\%).} Video retrieval performance under different K values on UCF101 and HMDB51. CACL$^{\ast}$ means pre-trained on Kinetics400 dataset}
\label{tab:recallatk}
\end{table*}

\begin{table*}[!h]
    \centering
   %\footnotesize
    \small
    {
    \begin{tabular}{p{3cm}ccccccc}
    % \begin{tabular}{pccccccc}
    \toprule
    Method&Year &input size&Pretrained&Architecture&Depth& UCF101 & HMDB51 \\
    \midrule
    % CoCLR-RGB~\cite{Han20} &NeurIPS'20& $32\times128$ & UCF101 & S3D&-& 81.4 & 52.1 \\
    % ViCC-RGB~\cite{toering2021self} &-& $32\times128$ & UCF101 & S3D&-& 84.3 & 47.9 \\
    % \midrule
    Puzzle~\cite{kim2019self} &AAAI'19& $16\times112$ & UCF101 & C3D&10& 65.0&31.3 \\
    VCOP~\cite{xu2019self} &CVPR'19& $16\times112$ & UCF101 & C3D &10& 65.6 & 28.4 \\
    VCP~\cite{luo2020video} &AAAI'20& $16\times112$ & UCF101 & C3D & 10&68.5&32.5  \\
    PRP \cite{Yao2020} &CVPR'20& $16\times112$&UCF101&C3D&10&69.1&34.5\\
    RSPNet \cite{chen2021rspnet} &AAAI'21& $16\times112$&Kinetics400&C3D&10&76.7&\textbf{44.6}\\
    DSM \cite{Wang2021EnhancingUV} &AAAI'21& $16\times112$&UCF101&C3D&10&70.3&40.5\\
    MoCo+BE \cite{wang2021removing} &CVPR'21& $16\times112$&UCF101&C3D&10&72.4&42.3\\
    \textbf{CACL} &-& $16\times112$ & UCF101 & C3D&10&  \textbf{77.2}& 43.5 \\
    \textbf{CACL} &-& $16\times112$ & Kinetics400 & C3D&10&  \textbf{77.5}& - \\
    \hline
    Puzzle~\cite{kim2019self} &AAAI'19& $16\times112$ & UCF101 & R3D&18& 65.8&33.7 \\
    VCOP \cite{xu2019self} &CVPR'19& $16\times112$ & UCF101 & R3D &10& 64.9 & 29.5 \\
    VCP \cite{luo2020video} &AAAI'20& $16\times112$ & UCF101 & R3D &10& 66.0&31.5  \\
    PRP \cite{Yao2020} &CVPR'20& $16\times112$&UCF101&R3D&10&66.5&29.7\\
    % MemDPC \cite{han2020memory} &ECCV'20& $40\times224$&UCF101&R3D&-&69.2&-\\
    IIC \cite{tao2020self} &ACMMM'20 & $16\times112$&UCF101&R3D&10&74.4&38.8\\
    RSPNet \cite{chen2021rspnet} &AAAI'21& $16\times112$&Kinetics400&R3D&18&74.3&41.8\\
    VideoMoCo \cite{Pan2021VideoMoCoCV} &CVPR'21& $16\times112$&Kinetics400&R3D&18&74.1&43.6\\
    % MoCo+BE \cite{wang2021removing} &CVPR'21& $16\times112$&UCF101&R3D&34&83.4&53.7\\
    \textbf{CACL} &-& $16\times112$ & UCF101 & R3D&10&  \textbf{77.5}& \textbf{43.8} \\
    \hline
    VCOP \cite{xu2019self} &CVPR'19& $16\times112$ & UCF101 & R(2+1)D &10& 72.4 & 30.9 \\
    VCP \cite{luo2020video} &AAAI'20& $16\times112$ & UCF101 & R(2+1)D &10& 66.3&32.2  \\
    PRP \cite{Yao2020} &CVPR'20& $16\times112$&UCF101&R(2+1)D&10&72.1&35.0\\
    Pace Pred \cite{wang2020self} &ECCV'20& $16\times112$&UCF101&R(2+1)D&10&75.9&35.9\\
    Pace Pred \cite{wang2020self} &ECCV'20& $16\times112$&Kinetics400&R(2+1)D&10&77.1&36.6\\
    RSPNet \cite{chen2021rspnet} &AAAI'21& $16\times112$&Kinetics400&R(2+1)D&10&81.1&44.6\\
    VideoMoCo \cite{Pan2021VideoMoCoCV} &CVPR'21& $16\times112$&Kinetics400&R(2+1)D&18&78.7&\textbf{49.2}\\
    % ViCC-RGB~\cite{toering2021self} &-& $32\times128$&UCF101&R(2+1)D&-&82.8 & 52.4\\
    \textbf{CACL} &-& $16\times112$ & UCF101 & R(2+1)D&10& \textbf{82.5}&48.8 \\
    % \hline
    % MoCo+BE \cite{wang2021removing} &CVPR'21& $16\times112$&UCF101&R3D&34&83.4&53.7\\
    \bottomrule
    \end{tabular}
    }
    \vspace{-3mm}
    \caption{The top-1 accuracy (\%) on UCF101 and HMDB51 dataset. The accuracy is computed by averaging over three splits.}
    \label{tab:sota_action_recognition_cmp}
    \vspace{-4mm}
\end{table*}

To evaluate our CACL on self-supervised video representation learning, we conduct video retrieval experiments on UCF101 to verify the effectiveness of each individual component developed in the CACL. We apply C3D as 3D CNN in this experiment. Results show that all the developed components make clear contributions.

\paragraph{Shuffle degree prediction.}
We investigate the capability of our shuffle degree prediction (SDP) on learning temporal information from a video, and compare it with recent VCOP \cite{xu2019self} and PRP~\cite{Yao2020} which were developed specifically for self-supervised temporal learning. Results are compared in Table \ref{tab:effectiveness_of_shuffle_degree}, where our SDP outperforms VCOP considerably and achieves comparable results with PRP. This verifies the effectiveness of our SDP by predicting an Edit distance between two videos. This enables it to identify more differences in temporal details, which naturally aggregate more meaningful temporal information. The result of SDP$^{\ast}$ which means SDP without uniform sampling, demonstrates the importance of uniform shuffle-degree sampling.

%We denote \textit{shuffle degree prediction} as SDP, Table \ref{tab:effectiveness_of_shuffle_degree} suggests that respectable performance can be attained by only using the shuffle degree prediction task, which demonstrate our assumption that shuffle degree prediction task can encourage model to learn transferable video representations and benefit downstream tasks. We simply apply MoCo to self-supervised video representation learning and report retrieval results in Table \ref{tab:effectiveness_of_shuffle_degree}, the results are better than shuffle degree shows the superiority of contrastive learning based self-supervised learning. We combine shuffle degree prediction task with MoCo, Table \ref{tab:effectiveness_of_shuffle_degree} shows a significant improvement over MoCo result, it proves that our shuffle degree prediction task can learn a heterogeneous video representation with contrastive learning method, appropriate collaborate these two method can benefit each other.

% \begin{figure}[t]
% % \fbox{\rule{0pt}{2in} \rule{0.98\linewidth}{0pt}}
% \centering
%   \includegraphics[width=0.98\linewidth]{img/retrieve_result.pdf}
% %  \vspace{-3mm}
%  \caption{Video retrieval results on UCF101, videos are represented by their central frames.}
% \label{fig:retrieve_result}
% % \vspace{-4mm}
% \end{figure}

We further perform video-level contrastive learning by simply implementing MoCo with C3D networks (referred as V-MoCo), which is able to learn spatio-temporal features for video-level representation. As shown in Table \ref{tab:effectiveness_of_shuffle_degree}, it obtains higher performance than SDP which focuses on learning temporal information from videos, demonstrating that video-level contrastive learning is able to learn stronger video-level representation in general. 
Interestingly, by integrating our SDP with V-MoCo, the performance can be further improved considerably, suggesting that the temporal representation learned by our SDP can compensate strongly to general video-level representation. Therefore, SDP can work collaboratively with general video representation learning architecture, and plays an important role by capturing temporal features like an optical flow branch used in the widely-applied two-stream design \cite{Simonyan2014}.      

\paragraph{Transformer video encoder.} In our video-level contrastive learning (V-MoCo), a key improvement is to use a video transformer encoder \textit{jointly} with 3D CNN encoder, which are able to generate more diverse yet meaningful positive pairs. Table \ref{tab:effectiveness_of_shuffle_degree} shows that our method achieves a large improvement by using the transformer encoder, demonstrating the effectiveness of our cross-architecture design with the transformer encoder. 
%
%As illustrated in Figure \ref{fig:learning_scheme}, we maximize the similarities of 4 positive pairs generated from each video clip, by using different data augmentations and different encoders. We denote different data augmentations as $q,k$, a transformer encoder as $T$, a 3D CNN encoder as $G$, then we can generate four feature representations $Gq, Gk, Tq,Tk$, for a video clip. 
%
As shown in Table \ref{tab:effectiveness_of_video_transformer}, ($Gq,Gk$) means positive pairs by 3D CNN with different data augmentations, which is equal to performing the original MoCo on videos with our SDP. Using all possible positive pairs ($Gq,Gk$),($Gq,Tq$),($Gq,Tk$),($Tq,Tk$) is a full implementation of our CACL. 
\textit{The performance can be improved gradually by adding more groups of the positive pairs.} This demonstrates that our video transformer encoder can provide more meaningful contrast information.

%We find a progressive improvement with more positive pairs added, which demonstrate that our video transformer encoder can provide meaningful contrast information and benefit video representation learning.

\paragraph{Positive pairs.} To further study the impact of diverse positive pairs to self-supervised contrastive learning, we compute average similarities of the positive pairs on UCF101 test split 1, as reported in Table \ref{tab:positive_pairs_analysis}. We can have a number of observations as follows: (1) by comparing ($Gq,Gk$) with ($Gq,Tq$), we found that a same video clip performed with different data augmentations can have a higher similarity than that computed by different encoders. This suggests that different encoders can generate positive pairs with more diversity, which are often meaningful and compensate to those generated by 3D CNN in contrastive learning.
Therefore, our cross-architecture can boost the performance of self-supervised contrastive learning; 
(2) as expected, ($Gq,Tk$) has the smallest similarity by using different data augmentations and different encoders, which may indicate the largest diversity and in turn has a larger improvement than that of ($Gq,Tq$), as shown in Table \ref{tab:effectiveness_of_video_transformer};
(3) by using a same encoder like ($Gq,Gk$) or ($Tq,Tk$), the transformer encoder has a higher similarity than that of 3D CNN. But it still obtains a considerable improvement by providing compensatory information to 3D CNN (shown in Table \ref{tab:effectiveness_of_video_transformer}). 

%we can find that with same data augmentation setting, the similarity of positive pairs generated from transformer is higher than the similarity of positive pairs get from 3D CNN, demonstrating the powerful ability of transformer for fitting the underline data distribution.

\subsection{Video Retrieval}
In this section, we evaluate CACL on video retrieval task. Following VCOP~\cite{xu2019self}, the evaluation is performed on the split 1 of UCF101 dataset. The  network  is  fixed as a  feature  extractor  after  performing pre-training  on  the split 1 training set of UCF101. We sample ten 16-frames clips from each video, and then perform feed-forward computation to generate features from the last pooling layer (p5).
We apply spatial max-pooling on each clip to get clip-level features, and perform average-pooling over 10 clips to obtain a video-level representation. We use videos from  the test set (query videos) to retrieve the videos from the training set by computing cosine similarities.  Once a video having a same category with the query video appears in Top-k retrieval results, we consider it as a correct Top-k hit. For a fair comparison, we apply the Top-k accuracy (k=1, 5, 10, 20, 50) as evaluation metrics. We report our results on UCF101 and HMDB51, and compare our method with recent self-supervised methods in Table \ref{tab:recallatk}.

Our method outperforms the state-of-the-art methods on all evaluation metrics by a large margin. For example, with C3D, our method obtains a 43.2\% top-1 accuracy on UCF101, which outperforms recent RSPNet~\cite{chen2021rspnet} by 7.2\%. Our CACL also has higher performance than other recent methods with different network architectures, demonstrating the strong generalization ability of our method.

\subsection{Action Recognition}

We further compare our method with recent self-supervised methods on action recognition in Table \ref{tab:sota_action_recognition_cmp}. For a fair comparison, we conduct experiments on three widely-used 3D CNN backbones applied in previous works. Detailed configurations such as \textit{input size}, \textit{pretrain dataset} and \textit{backbone architecture} are listed in Table \ref{tab:sota_action_recognition_cmp}. We report classification top-1 accuracy on UCF101 and HMDB51 datasets, by computing an average classification accuracy over 3 test splits. Our CACL with C3D and R3D can achieve the state-of-the-art results on both datasets. For C3D, our method outperforms recent DSM~\cite{Wang2021EnhancingUV} considerably by 6.9\% and 3\% on UCF101 and HMDB51 respectively. By comparing with VCOP, our method increases the accuracy largely from 65.6\% to 77.2\% on UCF101, and from 28.4\% to 43.5\% on HMDB51. For R3D and R(2+1)D networks pre-trained on UCF101, our method achieves 77.5\% and 82.5\% on UCF101, 43.8\% and 48.8\% on HMDB51, consistently outperforming RSPNet~\cite{chen2021rspnet} pre-trained on Kinetics400.

The performance on action recognition demonstrates the effectiveness of our method. CACL not only learns more meaningful spatio-temporal representations by introducing a transformer encoder into self-supervised contrastive learning, but also develops a new temporal prediction of Edit distance which allows it to aggregate rich temporal information that compensates to video-level spatio-temporal representation. Remarkably, UCF101 is a relatively small dataset compared to Kinetics400, CACL pre-trained on UCF101 even beats some recent methods pre-trained on Kinetics400, which indicates a great potential of our methods.

\subsection{Transformer Attention Visualization}

\begin{figure}[t]
\centering
  \rotatebox{0}{\includegraphics[scale=0.41]{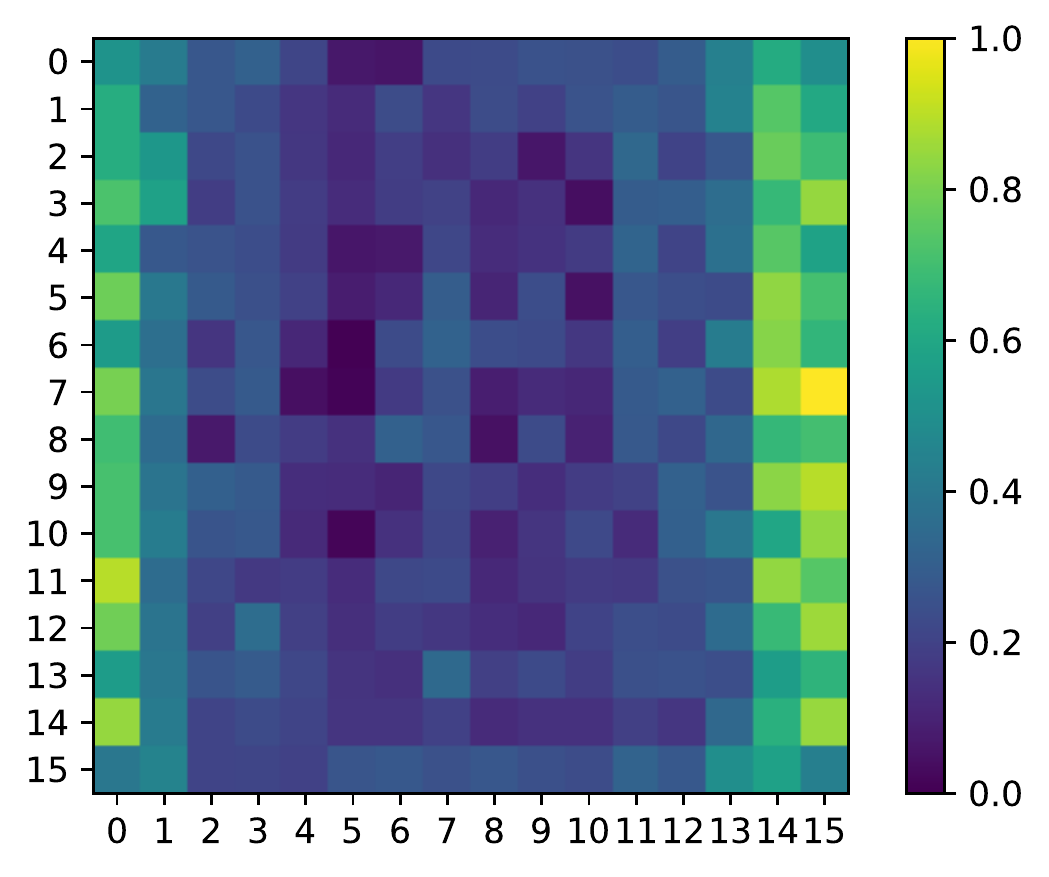}}
  \vspace{-3mm}
 \caption{Attention map of the first layer in the trained video transformer. 
 The numbers on both axis denote the indices of the tokens.
 Note that the attentions are normalized for better visualization.}
\label{fig:attn}
\vspace{-4mm}
\end{figure}

To justify the motivation of contrasting representations of CNNs and transformers, we study the attention range of the trained video transformer in CACL. 
We visualize the attention maps of the first layer in the video transformer, averaged over 50 randomly selected video clips. 
As shown in Figure \ref{fig:attn}, most frame tokens tend to pay more attention on the first and last positions than the middle ones of the clip. 
This is in line with our intuition that transformer can capture long-range dependencies, in contrast to the locality of 3DCNNs. Therefore, learning cross-architecture representations can compensate each other and achieve better performance.
%it may also suggests that the global feature is crucial for video understanding.

%---------------------------------------------------------------------------------
\section{Conclusion}

We have presented a new self-supervised video representation learning framework named CACL. We design a contrastive learning framework by introducing a transformer video encoder, which provide plentiful positive samples for 3D CNN in contrastive learning. We also introduce a new pretext which train a model to predict video shuffle degree. To verify the effectiveness of our approach, we conducted extensive experiments across three network architectures on two different downstream tasks. The experimental results indicates that our shuffle degree prediction and transformer video encoder  can encourage model to learn transferable video representations, the learned feature is heterogeneous with contrastive learning based method. 

\paragraph {\bf Acknowledgements.} {\small This work is supported by National Natural Science Foundation of China  (No.62076119, No.61921006).}

%%%%%%%%% REFERENCES
{\small
\bibliographystyle{ieee_fullname}
\bibliography{main}
}

\end{document}